\documentclass[11pt,a4paper]{article}
\usepackage{authblk}
\usepackage[hyperref]{acl2019}
\usepackage{times}
\usepackage{latexsym}
\usepackage{times}
\usepackage{latexsym}
\usepackage{natbib}
\usepackage{pslatex}
\usepackage[english]{babel}
\usepackage[utf8]{inputenc}
\usepackage{amsmath}
\usepackage{bm}
\usepackage{graphicx}
\usepackage{tikz}
\usepackage{tabu}
\usepackage{multirow}
\usepackage{url}
\usepackage{rotating}
\usepackage{natbib}
\usepackage{amssymb}
\usepackage{linguex}
\usepackage{tikz}
\usepackage{tikz-qtree}

\usepackage{url}

\newcommand{\key}{\textbf}

\aclfinalcopy 

\title{Hierarchical Representation in Neural Language Models: Suppression and Recovery of Expectations}

\author[1]{\textbf{Ethan Wilcox}}
\author[2]{\textbf{Roger Levy}}
\author[3]{\textbf{Richard Futrell}}

\affil[1]{Department of Linguistics, Harvard University, \tt{wilcoxeg@g.harvard.edu}}
\affil[2]{Department of Brain and Cognitive Sciences, MIT, \tt{rplevy@mit.edu}}
\affil[3]{Department of Language Science, UC Irvine, \tt{rfutrell@uci.edu}}

\date{}

\begin{document}

\setlength{\Exlabelwidth}{0.7em}
\setlength{\Exlabelsep}{0.7em}
\setlength{\SubExleftmargin}{1.3em}
\setlength{\Extopsep}{2pt}

\maketitle

\begin{abstract}

Deep learning sequence models have led to a marked increase in performance for a range of Natural Language Processing tasks, but it remains an open question whether they are able to induce proper hierarchical generalizations for representing natural language from linear input alone. 
Work using artificial languages as training input has shown that LSTMs are capable of inducing the stack-like data structures required to represent context-free and certain mildly context-sensitive languages \citep{weiss2018practical}---formal language classes which correspond in theory to the hierarchical structures of natural language. 
Here we present a suite of experiments probing whether neural language models trained on linguistic data induce these stack-like data structures and deploy them while incrementally predicting words.
We study two natural language phenomena: center embedding sentences and syntactic island constraints on the filler--gap dependency. 
In order to properly predict words in these structures, a model must be able to temporarily suppress certain expectations and then recover those expectations later, essentially pushing and popping these expectations on a stack. 
Our results provide evidence that models can successfully suppress and recover expectations in many cases, but do not fully recover their previous grammatical state.

\end{abstract}

\section{Introduction}

Deep learning sequence models such as RNNs \cite{elman1990finding, hochreiter1997long} have led to a marked increase in performance for a range of Natural Language Processing tasks \cite{jozefowicz2016exploring, dai2019transformer}, but it remains an open question whether they are able to induce hierarchical generalizations from linear input alone. Answering this question is important both for technical outcomes---models with explicit hierarchical structure show performance gains, at least when training on relatively small datasets \cite{choe-charniak:2016parsing,dyer2016rnng,kuncoro2016recurrent}---and for the scientific aim of understanding what biases, learning objectives and training regimes led to human-like linguistic knowledge. Previous work has approached this question by either examining models' internal state \cite{weiss2018practical,marevcek2018extracting} or by studying model behavior \cite{elman1991distributed,linzen2016assessing,futrell2019neural,mccoy2018revisiting}.

For this latter approach, much work has assessed sensitivity to hierarchy by examining whether the expectations associated with long-distance dependencies can be maintained even in the presence of intervening distractor words \cite{gulordava2018colorless,marvin2018targeted}. For example, \citet{linzen2016assessing} fed RNNs with the prefix \textit{The keys to the cabinet\dots}. If models assigned higher probability to the grammatical continuation \textit{are} over the ungrammatical continuation \textit{is}, they can be said to have learned the correct structural relationship between the subject and the verb, ignoring the syntactically-irrelevant singular distractor, \textit{the cabinet}. Work in this paradigm has uncovered a complex pattern in terms of what specific hierarchical structures are and are not represented by neural language models.

At the same time, work using artificial languages as input has demonstrated that LSTMs are capable of inducing the data structures required to produce hierarchically-structured sequences. For example, \citet{weiss2018practical} showed that LSTMs can learn to produce strings of the form $a^n b^n$, corresponding to context-free languages \citep{chomsky1956three}, and $a^n b^n c^n$, corresponding to mildly context-sensitive languages. Producing these strings requires a stack-like data structure where some number of $a$s are pushed onto the stack so that the same number of $b$s can be popped from it. The hierarchical structures of natural language are widely believed to be mildly context-sensitive \citep{shieber1985evidence,weir1988characterizing,seki1991multiple,joshi1997tree,kuhlmann2013mildly}, so this result shows that LSTMs are practically capable of inducing the proper data structures to handle the hierarchical structure of natural language.

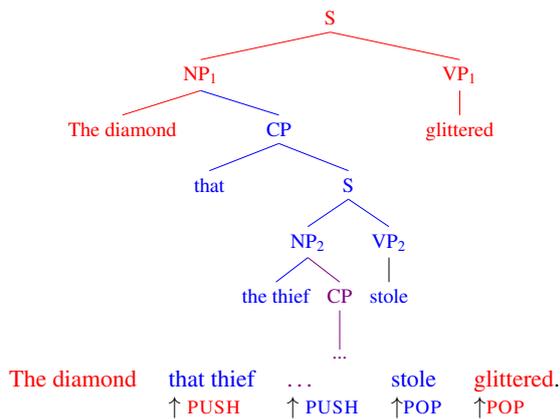
\begin{figure}
    \centering
    \scalebox{0.7}{
\begin{tikzpicture}
\Tree[.\textcolor{red}{S} 
        \edge[red]; [.\textcolor{red}{NP$_1$} \edge[red]; { \textcolor{red}{The diamond} }
            \edge[blue]; [.\textcolor{blue}{CP} \edge[blue]; {\textcolor{blue}{that}}
                \edge[blue]; [.\textcolor{blue}{S} 
                    \edge[blue]; [.\textcolor{blue}{NP$_2$} \edge[blue]; {\textcolor{blue}{the thief}} 
                        \edge[violet]; [.\textcolor{violet}{CP} \edge[violet]; {\textcolor{violet}{...}} ]
                    ]
                    \edge[blue]; [.\textcolor{blue}{VP$_2$} {\textcolor{blue}{stole}} ]
                ]
           ]
        ]
        \edge[red]; [.\textcolor{red}{VP$_1$} \edge[red]; {\textcolor{red}{glittered}} ]
    ]
    \end{tikzpicture} }
    \small{
    
    \begin{tabular}{lllll}
    \textcolor{red}{The diamond} & \textcolor{blue}{that thief} & \textcolor{violet}{\dots} &  \textcolor{blue}{stole} & \textcolor{red}{glittered}. \\
    & $\uparrow$ \textcolor{red}{\textsc{push}} & $\uparrow$ \textcolor{blue}{\textsc{push}} & $\uparrow$\textcolor{blue}{\textsc{pop}}  & $\uparrow$\textcolor{red}{\textsc{pop}} \\
    \end{tabular}
    }
    
\caption{Anatomy of a center embedding sentence. At each point marked \textsc{push}, comprehenders need to push the expectations generated by the subject noun onto a stack-like data structure, and suppress those expectations going forward. At the points marked \textsc{pop}, they must recover those expectations.}
    \label{fig:center-embed-schematic}
\end{figure}

What remains to be seen in a general way is that LSTMs induce and use these structures when trained on natural language input, rather than artificial language input. In this work, we present two suites of experiments that probe for evidence of hierarchical generalizations using two linguistic structures: \key{center embedding sentences} and \key{syntactic island constraints} on the filler--gap dependency. These structures exemplify context-free hierarchical structure in natural language. In order to correctly predict words in these structures, a model must use something like a stack data structure: certain expectations must be temporarily suppressed (pushed onto a stack), then recovered later at the right time and in the right order (popped from the stack in last-in-first-out order), as shown in Figure~\ref{fig:center-embed-schematic}.

For both of these contexts we assess how well RNNs can suppress local expectations within intervening blocking-structures and recover expectations on the far side. Success at these tasks would provide evidence that models not only ignore intervening material, but modulate and recover local expectations based on relative location within a syntactic structure.

\key{Center embeddings} are sentences in which a clause is embedded within the center of another clause, such that the expectations based on the external clause must be temporarily suppressed during the internal clause, and then recovered once the internal clause is complete. Such sentences were used as the original argument that natural language is not a regular language, but rather at least context-free \citep{chomsky1956three}. We find that neural language models can successfully suppress and recover expectations in sentences with two-layer embedding depth, but their accuracy depends on the particular lexical items used.

\key{Syntactic Islands} are structural configurations that block the filler--gap dependency, which is the dependency between a wh-word, such as \textit{who} or \textit{what}, and a gap, which is an empty syntactic position.  Using controlled experimental material, we find that models are able to suppress expectations for gaps inside two island constructions and partially recover them on the far side. However, the recovered expectation is far weaker than in non-island sentences and only robust in one of the models tested. Together, both experiments provide new evidence that RNN language models can approximate a soft notion of hierarchy to drive predictions, suppressing local expectations in some contexts and reactivating them based on relative syntactic position.

Overall our results show that the LSTMs tested have learned an approximate stack-like data structure to predict natural language, but the deployment of this structure depends on the particular lexical items used, and the recovery of expectations is often imperfect, especially for structures requiring deep stacks.

\section{Experimental Methodology}

In this work, we adapt psycholinguistic experimental techniques for neural model assessment. In this paradigm, neural models are fed hand-crafted sentences designed to belie underlying network knowledge. Following standard practice in psycholinguistics, statistical significance is derived from linear mixed-effects models \cite{baayen2008mixed}, with sum-coded fixed-effect predictors and maximal random slope structure \citep{barr2013random}.  This method permits us to factor out by-item variation and focus on differences in model behavior on materials differing only in the linguistic features of critical interest. \footnote{Our studies were preregistered on \url{aspredicted.org}: To see the preregistrations go to \url{http://aspredicted.org/blind.php?x=}$X$ where $X \in \{\texttt{uw873w}, \texttt{95gj46}$\}.}

\subsection{Neural Models Tested}

We study the behavior of two \textbf{LSTM} Language Models, one \textbf{Transformer} model and one baseline \textbf{N-gram} model, all trained on English text. The first LSTM is the ``BIG LSTM+CNN Inputs'' from \cite{jozefowicz2016exploring}, which we will refer to as the \textit{Google Model}. It was trained on the One Billion Word benchmark \cite{chelba2013one}, with two hidden layers of 8196 per layer and uses Convolutional Neural Net (CNN) character embeddings as input. The second LSTM model is the best-performing LSTM presented in the supplementary materials of \citet{gulordava2018colorless}, which we will refer to as the \textit{Gulordava Model}. It is much smaller, with 650 hidden units per layer, and was trained on 90-million words of Wikipedia. The Google model is current state-of-the art for an LSTM model unenriched with structural supervision, and the Gulordava model has been assessed extensively (e.g. \citealt{gulordava2018colorless,futrell2018rnns,wilcox2018rnn,giulianelli2018under}). The transformer model used here is the one presented in \citet{dai2019transformer}. It was trained on the Billion Word Benchmark and has 0.8 Billion parameters. The baseline is a 5-gram language model with Kneser-Ney smoothing, trained on the British National Corpus \cite{leech1992100} using SLIRM V1.5.7 \cite{stolcke2002srilm}.

\subsection{Dependent Measure: Surprisal}

We assess model behavior by measuring the \key{surprisal values} RNN language models assign to each word in a given sentence. Surprisal is the inverse log probability of a word given its context:

\begin{equation*}
S(x_i) = -\log_2 p(x_i|h_{i-1}),
\end{equation*}

In this case, $x_i$ is the current word and $h_{i-1}$ is the RNN's hidden state before processing $x_i$. The probability is calculated from the RNN's softmax layer, and the logarithm is taken in base 2 so that the surprisal is measured in bits. The surprisal at a certain word tells us the extent to which that word is expected under the language model's probability distribution. There is a strong tradition linking surprisal values derived from language models to psycholinguistic metrics, such as reading times in humans \cite{hale2001probabilistic, levy2008expectation, smith2013effect, goodkind2018predictive}.

\section{Center Embeddings}

\begin{figure*}
\centering
\begin{minipage}{0.98\textwidth}
\centering
\includegraphics[width=\textwidth]{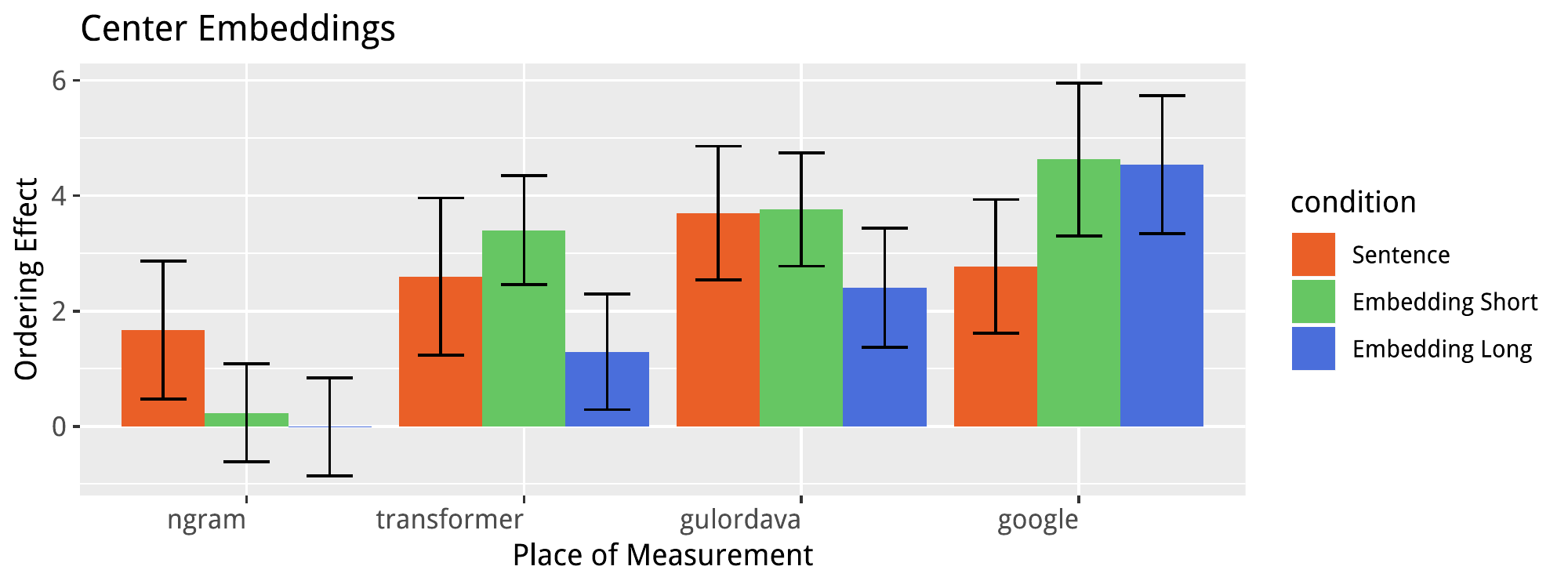}
\end{minipage}
\caption{Model results for center embedding sentences. Higher values indicate stronger divergence between the ordering effect \textit{match} and \textit{mismatch} conditions, indicating that models have learned the proper subject-verb pairings for the center embedding construction.}
\label{fig:center-embed}
\end{figure*}

In a center embedding sentence, the subject of a matrix (or main) clause is modified by an object-extracted relative clause. Because any Noun Phrase can serve as the host of a relative clause, the subject of the embedded relative clause can recursively serve as the start of a second center-embedding sentence, and so on \textit{ad infinitum}, provided that there are an equal number of subjects and verbs, as in Example~\ref{ex:long-center-embedding}.

\ex. The water [that the customer [that the waiter$_x$ disliked]$_y$ drank]$_z$ was cold. \label{ex:long-center-embedding}

Center embedding sentences exemplify the pattern $a^n b^n$, characteristic of context-free grammars, for natural language. However, the structure requires more than just counting: it is not sufficient that the number of verbs match the number of subjects, rather the \emph{verbs must semantically and syntactically match} their appropriate subjects and objects. The verb \textit{drank} is to be expected at the position marked $y$ in Example~\ref{ex:long-center-embedding}, but not at $x$ or $z$, because it corresponds to the subject \textit{customer} and the object \textit{water}. An incremental predictor must suppress an expectation for the word \textit{drink} during the region containing $x$, and then recover this expectation at $y$.

To assess whether the RNN LMs tested could suppress expectations for verbs set up by subjects and activate them in the correct order, we created 40 test items following the template in \ref{ex:center-embeddings}.

\ex. \label{ex:center-embeddings} 
\a. The diamond that the thief \textbf{stole$_{VP1}$ glittered$_{VP2}$}. [match, embedding] \label{ex:center-embed-gram}
\d. The diamond that the thief \textbf{glittered$_{VP1}$ stole$_{VP2}$}. [mismatch, embedding] \label{ex:center-embed-ungram}
\e. The diamond that the thief in the black mask \textbf{stole$_{VP1}$ glittered$_{VP2}$}. [match, embedding-long]
\f. The diamond that the thief in the black mask \textbf{glittered$_{VP1}$ stole$_{VP2}$}. [mismatch, embedding-long]
\b. The thief \textbf{stole$_{VP1}$} / The diamond \textbf{glittered$_{VP2}$} [match, sentence] \label{ex:center-embed-sentence-match}
\b. The thief \textbf{glittered$_{VP1}$} / The diamond \textbf{stole$_{VP2}$} [mismatch, sentence] \label{ex:center-embed-sentence-mismatch}
\z.

We use \textit{plausibility match} of ordering effect to assess whether the model was linking the right subject with the right verb. For example, it is plausible that a diamond glitters and a thief steals, as in \ref{ex:center-embed-gram}, but implausible that a thief glitters and a diamond steals as in \ref{ex:center-embed-ungram}.  In our test sentences the matrix clause subject tended to be an inanimate entity that took an intransitive verb, and the relative clause subject tended to be an animate entity that took a transitive verb.  For each item, we measure the strength of the models' expectation in terms of what we call the \key{ordering effect} at each verb: the surprisal in the [mismatch] condition minus the surprisal in the [match] condition. 
 Our prediction is that if a model has learned the ordering restrictions imposed by the grammatical rules that govern English center embedding and uses these restrictions to appropriately guide predictions about upcoming words, the ordering effect should be at least as great in the two [embedding] conditions as in the [sentence] conditions.  
 We report the summed ordering effect across the two VPs, which indicates the difference in surprisal between the two conditions due to specific order of the two verbs.
As control sentences, we converted each item into a pair of simple subject-verb sentences with no embedding, as in \ref{ex:center-embed-sentence-match}--\ref{ex:center-embed-sentence-mismatch}.  If the ordering effect for the control sentence conditions is not positive, it would call into question our selection of subject--verb pairs.

The results from this experiment can be seen in Figure \ref{fig:center-embed}, with the N-Gram model at left, the Transformer model center left and the two LSTM models at right. Error bars indicate 95\% confidence intervals of across item means, with within-item means subtracted, as advocated in \citet{masson2003using}. The baseline N-Gram model shows a positive ordering effect in the control \textit{Sentence} conditions, however the ordering effect is not significantly different from zero in the two \textit{Embedding} conditions. For the Transformer and two LSTM models, the ordering effect is positive in the control \textit{Sentence} conditions, as well as in the two critical \textit{Embedding} conditions. Examining the contributions of the individual items themselves, we find that the surprisal difference at the second (matrix) verb is responsible for the majority of the effect. That is, given the context \textit{The diamond that the thief ...} the continuations \textit{stole} and \textit{glittered} are equally likely. However, given the partially-saturated contexts in \ref{ex:context}, the continuation \textit{glittered} is much more likely in \ref{ex:context-gram} than the continuation \textit{stole} is in \ref{ex:context-ungram}.

\ex. \label{ex:context}
\a. The diamond that the thief stole... \label{ex:context-gram}
\b. The diamond that the thief glittered... \label{ex:context-ungram}

It is this difference that drives the majority of the Ordering Effect for the LSTM and Transformer models. Crucially, this behavior is inconsistent with a linear approach to subject/verb plausibility match. If the models had learned only that a semantically plausible verb needed to follow a subject, then the order of the verbs should have no effect on surprisal. The positive ordering effect we see in the two \textit{Embedding} conditions indicates the neural models have learned that the outer verb needs to be associated with the first subject: all three models exhibit a first-in-last-out approach to licensing consistent with stack-like representation.


Turning to differences between the three neural models: For the Gulordava and Transformer models the ordering effect is higher in the control \textit{Sentence} and \textit{Embedding Short} conditions than in the \textit{Embedding Long} conditions, although neither of the differences are significant. But for the Google model, the ordering effect is \emph{larger} in the embedding conditions than in the control sentence condition.  Although this increased effect size may at first glance be surprising, recall that in the embedding conditions, there is more preceding context than in  the control-sentence condition that is available to predict both verbs---including both arguments of the transitive verb.  This larger overall ordering effect in the embedding conditions suggests that the Google model, which is trained on an order of magnitude more data, may be more efficiently leveraging this additional preceding context.  It remains an open question why the Transformer Model, which is trained on the same large dataset, is unable to leverage similar contextual cues and maintain equally strong verbal expectations across the relative clause modifier.



\section{Filler--Gap Dependency Licensing}

\subsection{Measuring the Filler--Gap Dependency}

In English, a range of linguistic structures---such as questions and relative clauses---are formed by inserting a wh-word and eliding (or \textit{gapping}) subsequent material. For example, to turn the transitive sentence in \ref{ex:transitive} into a question, a filler ( \textit{who}) is inserted at the beginning of the clause, and the material being questioned (the direct object) is gapped, which we represent using the underscores (these are for presentational purposes only and are not included in test items). 

\ex. \label{ex:filler-gap}
\a. The count insulted the hostess yesterday. \label{ex:transitive}
\b. \textbf{Who} did the count insult \textbf{\_\_} yesterday? \label{ex:question}

Crucially, the filler and the gap depend on each other, insofar as a  filler word is illicit without a subsequent gap, and a gap is unlicensed without an upstream filler. \citet{wilcox2018rnn} established that the two LSTM language models tested here learn the filler--gap dependency insofar as they learn the 2 $\times$ 2 contingency between fillers and gaps. To assess this, for each of their test sentences they create four items following the four possible combinations of fillers and gaps, as in \ref{ex:wh-licensing} (note that in these and subsequent examples the * indicates ungrammatically).

\ex. \label{ex:wh-licensing}
\a. I know that the count insulted the hostess yesterday. [--FILLER, -GAP]
\b. * I know who the count insulted the hostess yesterday. [+FILLER, -GAP]
\c. * I know that the count insulted \_\_ yesterday. [--FILLER, +GAP]
\d. I know who the count insulted \_\_ yesterday. [+FILLER, +GAP]

Their logic is as follows: If the models are learning that gaps require fillers to be licensed, then the transition from an object-taking verb to a prepositional phrase that indicates a syntactic gap should be less surprising in the \textit{presence} of an upstream, licensing filler. That is S([--FILLER, +GAP]) should be greater than S([+FILLER, +GAP]) in the post gap material ``yesterday''. We refer to this difference as the \textit{+GAP wh-effect}, a large effect here indicates that the model has learned that gaps require fillers to be licensed. We measure the +GAP wh-effect in temporal adjuncts following the gap site, as in \textit{yesterday} in \ref{ex:wh-licensing}.

Additionally, if the models are learning that fillers set up expectations for gaps, then a filled argument structure position such as a direct object should be less surprising in the \textit{absence} of an upstream filler, a phenomena which is known in the psyhcolinguistics literature as the \key{filled gap effect}. That is, S([+FILLER, --GAP]) should be greater than S([--FILLER, --GAP]). We refer to this difference as a \textit{-GAP wh-effect}, a large effect here indicates that models have learned that fillers set up expectations for gaps. We measure the -GAP wh-effect in the embedded verb direct object, e.g. at ``the hostess'' in \ref{ex:wh-licensing}.

\citet{wilcox2018rnn} sum differences into a single metric, the \textbf{wh-licensing interaction}, which they measure in a post-gap temporal adjunct. In this work, we eschew the wh-licensing and look instead at the two wh-effects in the +GAP and -GAP conditions. We do this for two reasons: First, collapsing all four surprisal values obfuscates which part of the contingency the models learn. It may be the case that the vast majority of the licensing interaction comes from surprisal differences in just one of the two conditions, a fact which would be hard to observe by studying the full interaction. Second, if upstream fillers set up expectations for empty argument structure positions, then the \textit{filled gap effect} should be most noticeable on the object itself, not in a subsequent adjunct. Measuring the wh-effect separately for each condition allows us to take our measurement at the precise location where we would expect the effect to be the largest.

\subsection{Licensing Over Syntactic Islands} \label{sec:over-islands}

\begin{figure}
    \centering
\begin{tikzpicture}
\tikzset{level distance=18pt}
\tikzset{sibling distance=2pt}
\Tree [.{} \edge[roof]; {$\alpha$} [.{} 
[.\node(filler){filler}; ]
[.{} \edge[roof]; {$\beta$} [.\textbf{{X}}
\edge[roof]; 
{$\gamma$} \node(gap){\_\_}; \edge[roof]; {$\delta$}
] \edge[roof]; \node(gap2){\_\_};  ] \edge[roof]; {$\nu$} ] \edge[roof]; {$\theta$} ]

\draw[style=dashed] (filler.south) .. controls +(south:1.6) .. node {\large\color{red}{$\times$}} (gap.south); 
\draw[style=dashed] (filler.south west) .. controls +(south west:3) and +(south:2) .. node {\large \color{blue}{\checkmark}} (gap2.south) ; 
    \end{tikzpicture}
\caption{Island constraints and filling gaps across islands. If node \textbf{X} is an island, then a filler outside \textbf{X} cannot associate with a gap inside \textbf{X}, but it can associate with a filler on the far side of \textbf{X}.  For our analyses, successful learning of an island constraint implies that we should \emph{not} see wh-effects at the first part of the material $\delta$ immediately following the potential gap site, but we \emph{should} see wh-effects in $\nu$, following a licit gap site.}
    \label{fig:island-constraint-anatomy}
\end{figure}
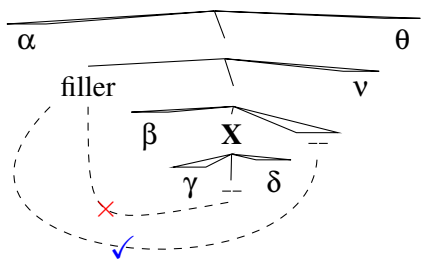

\begin{figure*}
\centering
\begin{minipage}{0.49\textwidth}
\centering
\includegraphics[width=\textwidth]{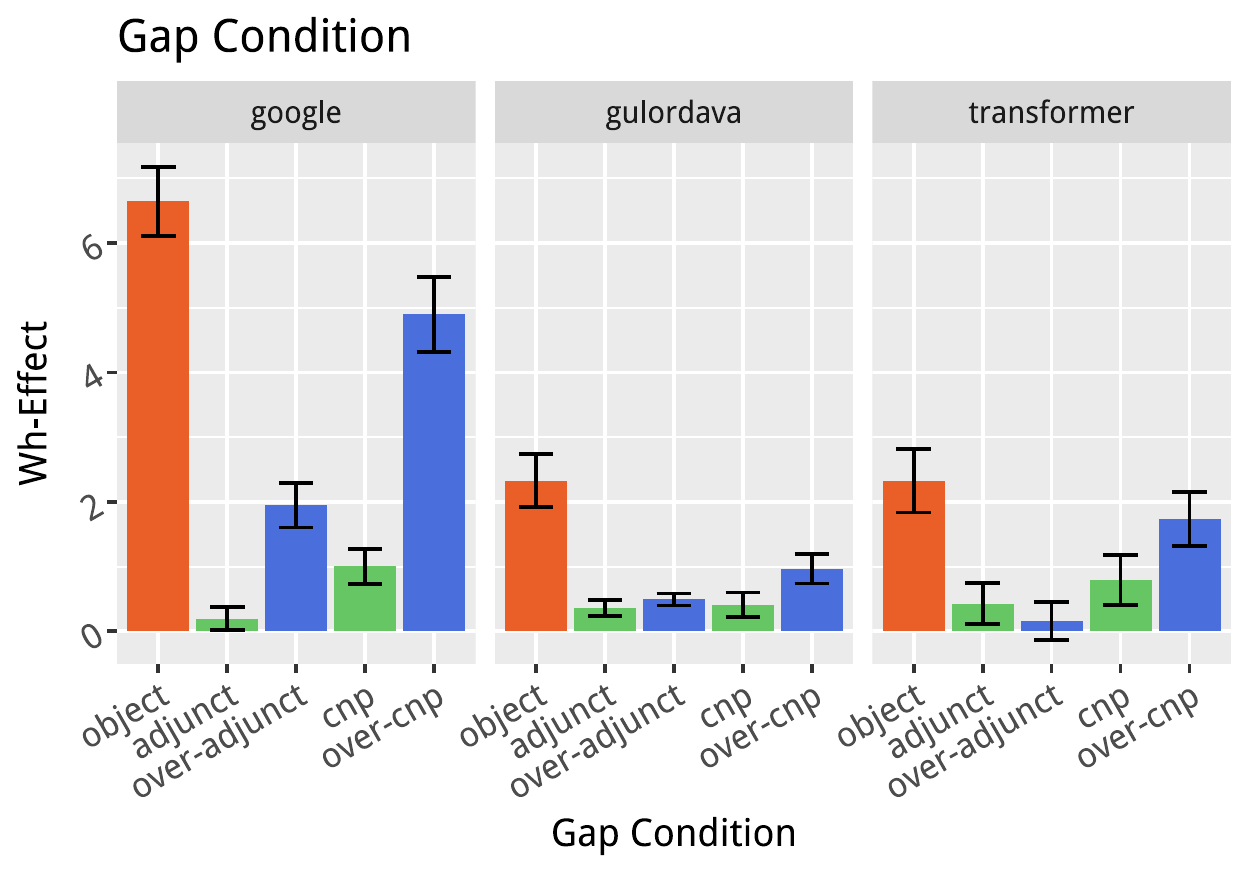}
\end{minipage}
\hspace{4pt}
\begin{minipage}{0.49\textwidth}
\centering
\includegraphics[width=\textwidth]{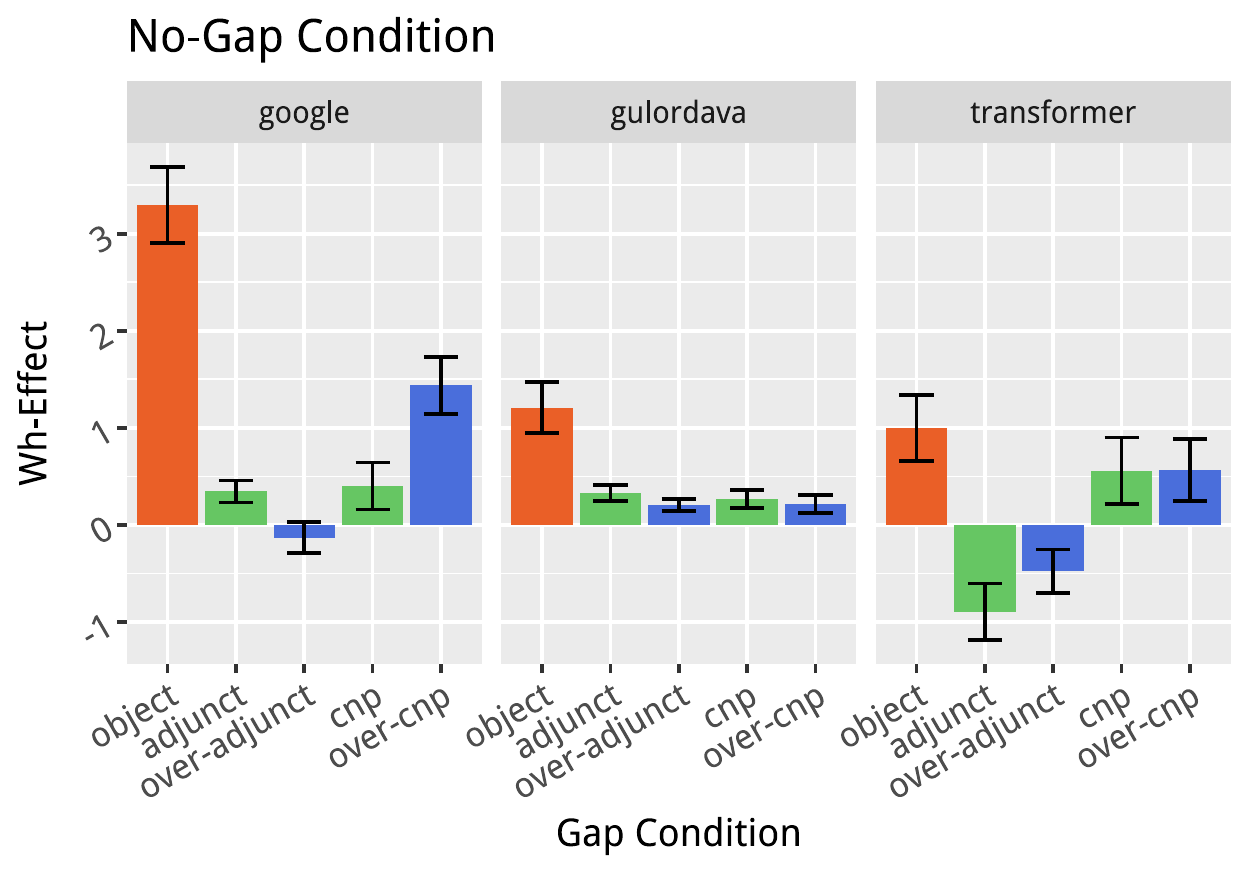}
\end{minipage}
\caption{Model results for maintaining the filler--gap dependency over island constructions. Strong wh-effects are expected in the grammatical  conditions (orange and blue), with reduced wh-effects in the island conditions (green). 
}
\label{fig:over-islands}
\end{figure*}

In addition to basic filler--gap dependency licensing, \citet{wilcox2018rnn} and \citet{wilcox2019syntactic} argue that the RNNs tested show sensitivity to numerous \key{island effects} (although see \citet{chowdhury2018rnn} for a contrasting view). Islands are syntactic positions that locally block the filler--gap dependency \cite{ross1967constraints}. For example, fillers can associate with gaps located in object position of a matrix clause, as in \ref{ex:island-good}, but not when the gap occurs within a relative clause, as in \ref{ex:island-bad}.

\ex. \label{ex:island}
\a. Who did the hostess insult \_\_ yesterday? \label{ex:island-good}
\b. * Who did the hostess insult [$_{\text{RC}}$ the count that knows \_\_ ] yesterday? \label{ex:island-bad} 

Crucially, although islands block the fillers from associating with gaps within the island, they do not prohibit association between fillers and gaps that occur structurally to the right of the island, as shown in Figure~\ref{fig:island-constraint-anatomy}. 

\citet{wilcox2019structural} found that while large scale models are able to thread the 2 $\times$ 2 contingency between fillers and gaps into syntactically complex material--such as through numerous sentential embeddings---they do not thread the dependency into some island configurations. Inside of relative clauses and temporal adjuncts, for example, the presence or absence of an upstream filler has no effect on the relative surprisal of a gap, and the wh-licensing interaction drops to near zero.

However, model inability to thread the filler--gap dependency into island configurations provides only half of the evidence necessary to establish that neural models are ``learning" islands in a way meaningfully similar to humans. Island configurations act as blockers, but only for the duration of the island---the length of the relative clause or the temporal adjunct, for the two islands tested here. If RNNs learn islands as local contexts into which an outside filler cannot license a gap, they should recover their expectations for gaps following the island. 

To assess whether models recover expectations for licit gaps following island configurations, we generated test sentences following the template in \ref{ex:over-gap}, featuring two well-studied islands: \textbf{adjunct islands} \ref{ex:adjunct} and \textbf{complex noun phrase islands} \ref{ex:cnp}. In these examples, the island portions of the sentences, in which gaps are not allowed, appear in boldface.

\ex. \label{ex:over-gap} 
\a. I know who the count from the southern province talked very loudly with \_\_ on the balcony. [object] \label{ex:object}
\b. * I know who , \textbf{after the count insulted  \_\_ on the balcony} , the hostess talked with the countess. [adjunct] \label{ex:adjunct}
\c. I know who , \textbf{after insulting the hostess }, the count talked with  \_\_ on the balcony. [over-adjunct] \label{ex:over-adjunct}
\d. * I know who \textbf{the count that insulted  \_\_ on the balcony} talked with the hostess. [cnp] \label{ex:cnp} \label{ex:cnp}
\e. I know who \textbf{the count that insulted the hostess} talked loudly with  \_\_ on the balcony. [over-cnp] \label{ex:over-cnp}

For each condition, we created a sentence template and seeded each region in the template with between three and seven examples. Permuting the examples, we generated thousands of candidate sentences, from which we sampled 100 at random and measured the wh-effect for the +GAP and {--GAP} conditions. If the models are sensitive to the island constraints, then we expect strong wh-effects in the grammatical [object] condition, but not in the ungrammatical [adjunct] and [complex noun phrase] ([cnp]) conditions. Furthermore, if models are able to recover expectations from gaps following the end of an island, we would expect strong wh-effects in the grammatical [over-adjunct] and [over-cnp] conditions. 

The results from this experiment can be seen in Figure \ref{fig:over-islands}, with the wh-effect in the +GAP condition at left and the --GAP condition at right.  The baseline N-Gram model showed wh-effects that were not significantly different from zero for all conditions, and is not included in the graphs.
Focusing on the +GAP condition at left, we see a strong wh-effect in the control \textit{object} condition but a significant reduction of wh-effect in the \textit{adjunct} and \textit{cnp} conditions for all models ($p<0.001$). In the grammatical \textit{over-adjunct} and \textit{over-cnp} we still see a significant reduction in wh-effect compared to the \textit{object} condition ($p<0.001$), but a significant increase in wh-effect relative to the corresponding island conditions in many cases. This recovery of expectations is significant for CNP Islands for all models ($p<0.001$) and for the Adjunct Islands in the case of the Google model ($p<0.001$). The results are especially striking for the Google Model: While the absence of an upstream filler induces only one more bit of surprisal at the gap site within an island, it induces between 2-5 more bits of surprisal when a gap occurs licitly downstream of an island.

Turning to the -GAP conditions at right, the results are more mixed. All three models show significantly more licensing interaction in the control \textit{object} condition compared to the island conditions, except for the Transformer model in the case of CNP Islands. However, only the Google Model shows a significant recuperation of empty argument structure expectation in the \textit{cnp} vs. \textit{over-cnp} condition ($p<0.001$). These results indicate that the three language models tested are able to bracket their expectations for gaps and regain them on the other side in the case of relative clauses. However, neither model does a good job of recovering the filled gap effect following an island, modulo complex noun phrase islands for the Google model.

\subsection{Wh-Discharge Effects}

The filler--gap dependency is constrained, insofar as fillers can license only one gap. \citet{wilcox2018rnn} found that RNN models were sensitive to this constraint, displaying a reduction in licensing interaction following a gap, if another gap existed upstream in the sentence as in \ref{ex:double-gap-simple}. The presence of a filler sets up an expectation for a gap, which is discharged at the first gap site, and cannot participate in downstream licensing effects. However, if models are sensitive to the fact that gaps cannot licitly occur within islands (unless they are licensed within the island itself), the presence of a gap inside a relative clause or a temporal adjunct should not result in the discharge of gap expectation. 

To assess whether gap discharge effects are mitigated when the first gap occurs inside of an island, we generated 100 examples following the process described in Section \ref{sec:over-islands} and the template in \ref{ex:double-gap}. Following the results in \citet{wilcox2018rnn}, section 3.3, we expect a slightly negative wh-effects in the \textit{subject} condition. However, if gaps inside of islands do not discharge the wh-effect set up by a filler, we expect positive wh-effects in the \textit{adjunct-discharge} and \textit{cnp-discharge} conditions.

\ex. \label{ex:double-gap} 
\a. I know who \_\_ talked very loudly with \_\_ on the balcony.
 [subject] \label{ex:double-gap-simple}
\b. I know who , after insulting \_\_ , the count talked loudly with \_\_ on the balcony. [adjunct-discharge]
\c. I know who the old man that insulted \_\_ talked loudly with \_\_ on the balcony. [cnp-discharge]

\begin{figure}
\centering
\begin{minipage}{0.49\textwidth}
\centering
\includegraphics[width=\textwidth]{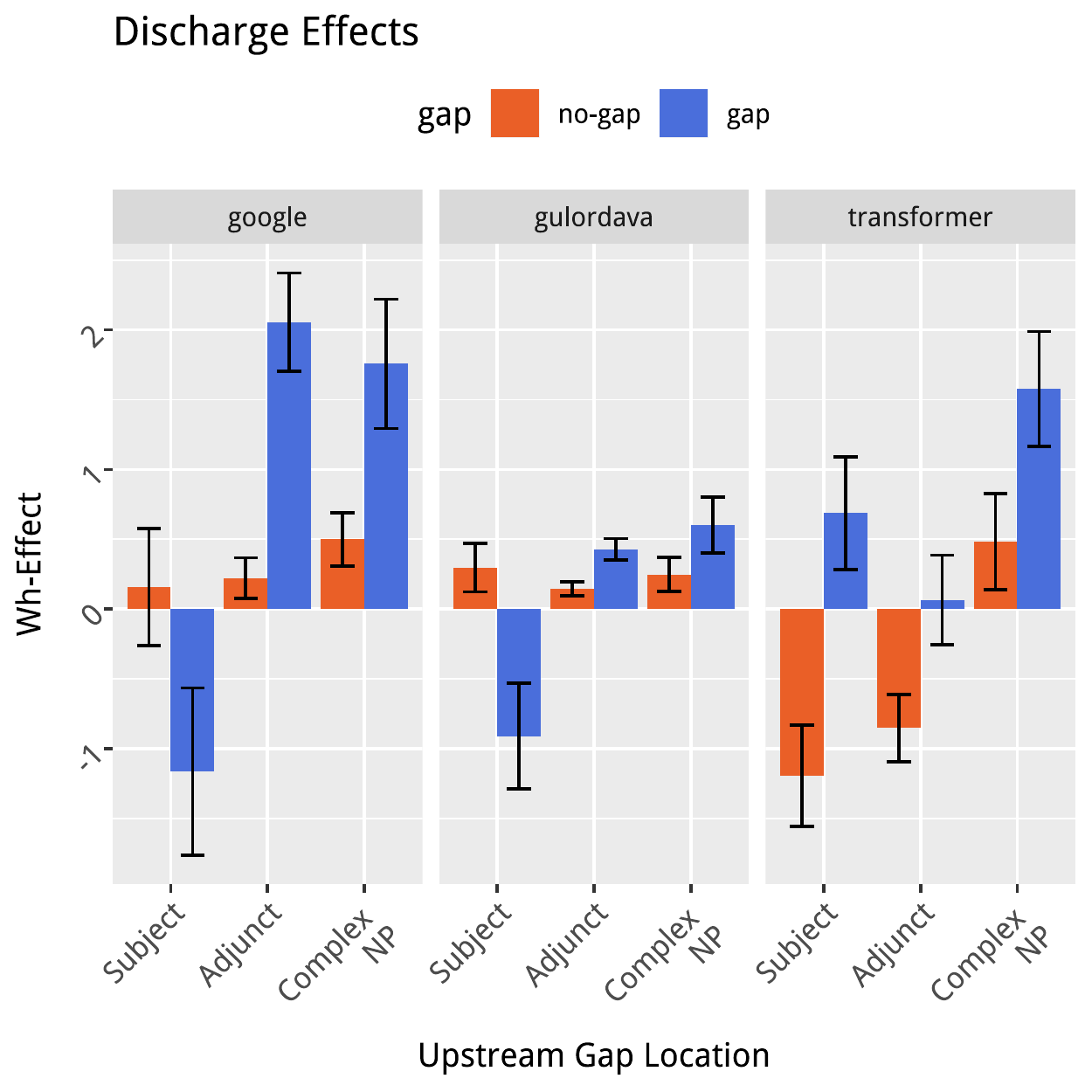}
\end{minipage}
\caption{Discharge effects for gaps in Subject and Island positions. Strong Wh-Effects are expected in the Adjunct and ComplexNP conditions, with negative wh-effects in the Subject condition.}
\label{fig:discharge}
\end{figure}

The results from this experiment can be seen in \ref{fig:discharge}. For the RNN models, In the --GAP cases, for both models there is no significant difference between the conditions. However, in the +GAP cases, there is a significant increase in wh-effect between the \textit{subject} and \textit{adjunct-discharge} and \textit{cnp-discharge} conditions ($p<0.001$ for both models). For the Transformer model, the \textit{Adjunct} and \textit{Subject} conditions pattern together, and there is a significant increase in Wh-Effect for the \textit{Complex NP} condition, in both the +Gap and -Gap cases ($p<0.001$). 

These results conform to those found in \ref{sec:over-islands}: all models have a difficult time threading expectations for filled argument structure positions through syntactically-complex material. However, expectations surrounding gaps are clear, at least for the two LSTM models: When gaps occur inside of islands, they do not trigger the the same discharge effects as gaps in subject positions. Interestingly, this generalization seems to be less robust for the Transformer model, which demonstrates the correct behavior only for Complex NP islands. Over all, these results provide further evidence that the models are able to process the edge of a syntactic island, and recover expectations for gaps on the far side. 

\section{General Discussion and Conclusion}

In this paper, we have provided new evidence that neural models can learn hierarchical generalizations from linear input alone. By adopting the psycholinguistic paradigm for RNN assessment, we have shown that two large-scale LSTM models and one Transformer modal can suppress and recover expectations set up by subject \textit{Noun Phrases} and \textit{fillers} within intervening blocking structures and recover those expectations on the far side of those syntactic blockers. This behavior corresponds to the idea of pushing and popping expectations in a stack-like data structure, which is required for proper incremental prediction of context-free languages. 

However, the suppression and recovery of expectations is imperfect. 
For example, in the filler--gap dependency, we found that models only partially recover expectations for gaps on the far side of island structures, especially in the -GAP conditions, where no model was able to robustly recover \textit{filled gap} expectations. Interestingly, the LSTM models tended to perform better than Transformer model, even when  trained on orders of magnitude less data. These results indicate that the large number of parameters in the Transformer architecture may result in lower test-time perplexity, but may not necessarily result in more grammatical behavior, at least for the tightly-controlled syntactic test suites presented here. It may be that the smaller number of  parameters in the LSTMs force the models to make more robust, and ultimately humanlike generalizations.

This work only assesses two model architectures. It is likely that neural models with a stronger structural bias, such as RNNGs \cite{dyer2016rnng} or LSTMs enhanced with a structural bias as in \citet{shen2018ordered} would perform better on the tests presented here; testing these, and other models, will be the basis for future work.

\bibliography{acl2019}

\begin{thebibliography}{39}
\expandafter\ifx\csname natexlab\endcsname\relax\def\natexlab#1{#1}\fi

\bibitem[{Baayen et~al.(2008)Baayen, Davidson, and Bates}]{baayen2008mixed}
R~Harald Baayen, Douglas~J Davidson, and Douglas~M Bates. 2008.
\newblock Mixed-effects modeling with crossed random effects for subjects and
  items.
\newblock \emph{Journal of Memory and Language}, 59(4):390--412.

\bibitem[{Barr et~al.(2013)Barr, Levy, Scheepers, and Tily}]{barr2013random}
Dale~J Barr, Roger Levy, Christoph Scheepers, and Harry~J Tily. 2013.
\newblock Random effects structure for confirmatory hypothesis testing: Keep it
  maximal.
\newblock \emph{Journal of Memory and Language}, 68(3):255--278.

\bibitem[{Chelba et~al.(2013)Chelba, Mikolov, Schuster, Ge, Brants, Koehn, and
  Robinson}]{chelba2013one}
Ciprian Chelba, Tomas Mikolov, Mike Schuster, Qi~Ge, Thorsten Brants, Phillipp
  Koehn, and Tony Robinson. 2013.
\newblock One billion word benchmark for measuring progress in statistical
  language modeling.
\newblock \emph{arXiv preprint arXiv:1312.3005}.

\bibitem[{Choe and Charniak(2016)}]{choe-charniak:2016parsing}
Doo~Kok Choe and Eugene Charniak. 2016.
\newblock Parsing as language modeling.
\newblock In \emph{Proceedings of the 2016 Conference on Empirical Methods in
  Natural Language Processing}, pages 2331--2336.

\bibitem[{Chomsky(1956)}]{chomsky1956three}
Noam Chomsky. 1956.
\newblock Three models for the description of language.
\newblock \emph{IRE Transactions on Information Theory}, 2(3):113--124.

\bibitem[{Chowdhury and Zamparelli(2018)}]{chowdhury2018rnn}
Shammur~Absar Chowdhury and Roberto Zamparelli. 2018.
\newblock Rnn simulations of grammaticality judgments on long-distance
  dependencies.
\newblock In \emph{Proceedings of the 27th International Conference on
  Computational Linguistics}, pages 133--144.

\bibitem[{Dai et~al.(2019)Dai, Yang, Yang, Cohen, Carbonell, Le, and
  Salakhutdinov}]{dai2019transformer}
Zihang Dai, Zhilin Yang, Yiming Yang, William~W Cohen, Jaime Carbonell, Quoc~V
  Le, and Ruslan Salakhutdinov. 2019.
\newblock Transformer-xl: Attentive language models beyond a fixed-length
  context.
\newblock \emph{arXiv preprint arXiv:1901.02860}.

\bibitem[{Dyer et~al.(2016)Dyer, Kuncoro, Ballesteros, and
  Smith}]{dyer2016rnng}
Chris Dyer, Adhiguna Kuncoro, Miguel Ballesteros, and Noah~A. Smith. 2016.
\newblock Recurrent neural network grammars.
\newblock In \emph{Proc. of NAACL}.

\bibitem[{Elman(1990)}]{elman1990finding}
Jeffrey~L Elman. 1990.
\newblock Finding structure in time.
\newblock \emph{Cognitive Science}, 14(2):179--211.

\bibitem[{Elman(1991)}]{elman1991distributed}
Jeffrey~L Elman. 1991.
\newblock Distributed representations, simple recurrent networks, and
  grammatical structure.
\newblock \emph{Machine learning}, 7(2-3):195--225.

\bibitem[{Futrell et~al.(2018)Futrell, Wilcox, Morita, and
  Levy}]{futrell2018rnns}
Richard Futrell, Ethan Wilcox, Takashi Morita, and Roger Levy. 2018.
\newblock Rnns as psycholinguistic subjects: Syntactic state and grammatical
  dependency.
\newblock \emph{arXiv preprint arXiv:1809.01329}.

\bibitem[{Futrell et~al.(2019)Futrell, Wilcox, Morita, Qian, Ballesteros, and
  Levy}]{futrell2019neural}
Richard Futrell, Ethan Wilcox, Takashi Morita, Peng Qian, Miguel Ballesteros,
  and Roger Levy. 2019.
\newblock Neural language models as psycholinguistic subjects: Representations
  of syntactic state.
\newblock \emph{arXiv preprint arXiv:1903.03260}.

\bibitem[{Giulianelli et~al.(2018)Giulianelli, Harding, Mohnert, Hupkes, and
  Zuidema}]{giulianelli2018under}
Mario Giulianelli, Jack Harding, Florian Mohnert, Dieuwke Hupkes, and Willem
  Zuidema. 2018.
\newblock Under the hood: Using diagnostic classifiers to investigate and
  improve how language models track agreement information.
\newblock \emph{arXiv preprint arXiv:1808.08079}.

\bibitem[{Goodkind and Bicknell(2018)}]{goodkind2018predictive}
Adam Goodkind and Klinton Bicknell. 2018.
\newblock Predictive power of word surprisal for reading times is a linear
  function of language model quality.
\newblock In \emph{Proceedings of the 8th Workshop on Cognitive Modeling and
  Computational Linguistics (CMCL 2018)}, pages 10--18.

\bibitem[{Gulordava et~al.(2018)Gulordava, Bojanowski, Grave, Linzen, and
  Baroni}]{gulordava2018colorless}
Kristina Gulordava, Piotr Bojanowski, Edouard Grave, Tal Linzen, and Marco
  Baroni. 2018.
\newblock Colorless green recurrent networks dream hierarchically.
\newblock \emph{arXiv preprint arXiv:1803.11138}.

\bibitem[{Hale(2001)}]{hale2001probabilistic}
John Hale. 2001.
\newblock A probabilistic earley parser as a psycholinguistic model.
\newblock In \emph{Proceedings of the second meeting of the North American
  Chapter of the Association for Computational Linguistics on Language
  technologies}, pages 1--8. Association for Computational Linguistics.

\bibitem[{Hochreiter and Schmidhuber(1997)}]{hochreiter1997long}
Sepp Hochreiter and J{\"u}rgen Schmidhuber. 1997.
\newblock Long short-term memory.
\newblock \emph{Neural Computation}, 9(8):1735--1780.

\bibitem[{Joshi and Schabes(1997)}]{joshi1997tree}
Aravind~K Joshi and Yves Schabes. 1997.
\newblock Tree-adjoining grammars.
\newblock In \emph{Handbook of formal languages}, pages 69--123. Springer.

\bibitem[{Jozefowicz et~al.(2016)Jozefowicz, Vinyals, Schuster, Shazeer, and
  Wu}]{jozefowicz2016exploring}
Rafal Jozefowicz, Oriol Vinyals, Mike Schuster, Noam Shazeer, and Yonghui Wu.
  2016.
\newblock Exploring the limits of language modeling.
\newblock \emph{arXiv}, 1602.02410.

\bibitem[{Kuhlmann(2013)}]{kuhlmann2013mildly}
Marco Kuhlmann. 2013.
\newblock Mildly non-projective dependency grammar.
\newblock \emph{Computational Linguistics}, 39(2):355--387.

\bibitem[{Kuncoro et~al.(2016)Kuncoro, Ballesteros, Kong, Dyer, Neubig, and
  Smith}]{kuncoro2016recurrent}
Adhiguna Kuncoro, Miguel Ballesteros, Lingpeng Kong, Chris Dyer, Graham Neubig,
  and Noah~A Smith. 2016.
\newblock What do recurrent neural network grammars learn about syntax?
\newblock \emph{arXiv preprint arXiv:1611.05774}.

\bibitem[{Leech(1992)}]{leech1992100}
Geoffrey~Neil Leech. 1992.
\newblock 100 million words of english: the british national corpus (bnc).

\bibitem[{Levy(2008)}]{levy2008expectation}
Roger Levy. 2008.
\newblock Expectation-based syntactic comprehension.
\newblock \emph{Cognition}, 106(3):1126--1177.

\bibitem[{Linzen et~al.(2016)Linzen, Dupoux, and
  Goldberg}]{linzen2016assessing}
Tal Linzen, Emmanuel Dupoux, and Yoav Goldberg. 2016.
\newblock Assessing the ability of lstms to learn syntax-sensitive
  dependencies.
\newblock \emph{Transactions of the Association for Computational Linguistics},
  4:521--535.

\bibitem[{Mare{\v{c}}ek and Rosa(2018)}]{marevcek2018extracting}
David Mare{\v{c}}ek and Rudolf Rosa. 2018.
\newblock Extracting syntactic trees from transformer encoder self-attentions.
\newblock In \emph{Proceedings of the 2018 EMNLP Workshop BlackboxNLP:
  Analyzing and Interpreting Neural Networks for NLP}, pages 347--349.

\bibitem[{Marvin and Linzen(2018)}]{marvin2018targeted}
Rebecca Marvin and Tal Linzen. 2018.
\newblock Targeted syntactic evaluation of language models.
\newblock \emph{arXiv preprint arXiv:1808.09031}.

\bibitem[{Masson and Loftus(2003)}]{masson2003using}
Michael~EJ Masson and Geoffrey~R Loftus. 2003.
\newblock Using confidence intervals for graphically based data interpretation.
\newblock \emph{Canadian Journal of Experimental Psychology/Revue canadienne de
  psychologie exp{\'e}rimentale}, 57(3):203.

\bibitem[{McCoy et~al.(2018)McCoy, Frank, and Linzen}]{mccoy2018revisiting}
R~Thomas McCoy, Robert Frank, and Tal Linzen. 2018.
\newblock Revisiting the poverty of the stimulus: hierarchical generalization
  without a hierarchical bias in recurrent neural networks.
\newblock \emph{arXiv preprint arXiv:1802.09091}.

\bibitem[{Ross(1967)}]{ross1967constraints}
John~Robert Ross. 1967.
\newblock Constraints on variables in syntax.

\bibitem[{Seki et~al.(1991)Seki, Matsumura, Fujii, and
  Kasami}]{seki1991multiple}
Hiroyuki Seki, Takashi Matsumura, Mamoru Fujii, and Tadao Kasami. 1991.
\newblock On multiple context-free grammars.
\newblock \emph{Theoretical Computer Science}, 88(2):191--229.

\bibitem[{Shen et~al.(2018)Shen, Tan, Sordoni, and Courville}]{shen2018ordered}
Yikang Shen, Shawn Tan, Alessandro Sordoni, and Aaron Courville. 2018.
\newblock Ordered neurons: Integrating tree structures into recurrent neural
  networks.
\newblock \emph{arXiv preprint arXiv:1810.09536}.

\bibitem[{Shieber(1985)}]{shieber1985evidence}
Stuart~M Shieber. 1985.
\newblock Evidence against the context-freeness of natural language.
\newblock In \emph{Philosophy, Language, and Artificial Intelligence}, pages
  79--89. Springer.

\bibitem[{Smith and Levy(2013)}]{smith2013effect}
Nathaniel~J Smith and Roger Levy. 2013.
\newblock The effect of word predictability on reading time is logarithmic.
\newblock \emph{Cognition}, 128(3):302--319.

\bibitem[{Stolcke(2002)}]{stolcke2002srilm}
Andreas Stolcke. 2002.
\newblock Srilm-an extensible language modeling toolkit.
\newblock In \emph{Seventh international conference on spoken language
  processing}.

\bibitem[{Weir(1988)}]{weir1988characterizing}
David~Jeremy Weir. 1988.
\newblock Characterizing mildly context-sensitive grammar formalisms.

\bibitem[{Weiss et~al.(2018)Weiss, Goldberg, and Yahav}]{weiss2018practical}
Gail Weiss, Yoav Goldberg, and Eran Yahav. 2018.
\newblock On the practical computational power of finite precision rnns for
  language recognition.
\newblock \emph{arXiv preprint arXiv:1805.04908}.

\bibitem[{Wilcox et~al.(2019{\natexlab{a}})Wilcox, Levy, and
  Futrell}]{wilcox2019syntactic}
Ethan Wilcox, Roger Levy, and Richard Futrell. 2019{\natexlab{a}}.
\newblock What syntactic structures block dependencies in rnn language models?
\newblock \emph{arXiv preprint arXiv:1905.10431}.

\bibitem[{Wilcox et~al.(2018)Wilcox, Levy, Morita, and Futrell}]{wilcox2018rnn}
Ethan Wilcox, Roger Levy, Takashi Morita, and Richard Futrell. 2018.
\newblock What do rnn language models learn about filler-gap dependencies?
\newblock \emph{arXiv preprint arXiv:1809.00042}.

\bibitem[{Wilcox et~al.(2019{\natexlab{b}})Wilcox, Qian, Futrell, Ballesteros,
  and Levy}]{wilcox2019structural}
Ethan Wilcox, Peng Qian, Richard Futrell, Miguel Ballesteros, and Roger Levy.
  2019{\natexlab{b}}.
\newblock Structural supervision improves learning of non-local grammatical
  dependencies.
\newblock \emph{arXiv preprint arXiv:1903.00943}.

\end{thebibliography}
\bibliographystyle{acl_natbib}

\end{document}